\documentclass[journal,two side,web]{ieeecolor}
\usepackage{tmi}
\usepackage{cite}
\usepackage{amsmath,amssymb,amsfonts}
\usepackage{algorithmic}
\usepackage{graphicx}
\usepackage{textcomp}
\usepackage{subcaption}
\usepackage{url}
\usepackage{booktabs}
\newcommand{\citep}{\cite}
\usepackage[export]{adjustbox}
\usepackage{hyperref}
\usepackage{pifont}
\usepackage{colortbl}
\usepackage{color,soul}
\definecolor{mygray}{gray}{.9}

\def\BibTeX{{\rm B\kern-.05em{\sc i\kern-.025em b}\kern-.08em
    T\kern-.1667em\lower.7ex\hbox{E}\kern-.125emX}}
\begin{document}
\title{Interpretable Multimodal Learning for Cardiovascular Hemodynamics Assessment}

\author{Prasun C Tripathi, Sina Tabakhi,  Mohammod N I Suvon,  Lawrence Sch\"ob, Samer Alabed,  Andrew J Swift, Shuo Zhou, and Haiping Lu, \IEEEmembership{Senior Member, IEEE}
\thanks{The study was supported by the Wellcome Trust grants (215799/Z/19/Z and 205188/Z/16/Z) and National Institute for Health and Care Research (NIHR), Sheffield Biomedical Research Centre (NIHR203321).}
\thanks{Prasun C Tripathi, Sina Tabakhi, Mohammod N. I. Suvon, Shuo Zhou, Lawrence Sch\"ob, and Haiping Lu are with Centre for Machine Intelligence (CMI) and the Department of Computer Science,  University of Sheffield, S1 4DP, U.K. (Corresponding Author: Haiping Lu, E-mail: h.lu@sheffield.ac.uk). }
\thanks{Samer Alabed and Andrew J Swift are with School of Medicine and Population Health and Institute for in Silico Medicine (INSIGNEO), University of Sheffield, S10 2TA, U.K.}
}

\maketitle

\begin{abstract}

Pulmonary Arterial Wedge Pressure (PAWP) is an essential cardiovascular hemodynamics marker to detect heart failure. In clinical practice, Right Heart Catheterization is considered a gold standard for assessing cardiac hemodynamics while non-invasive methods are often needed to screen high-risk patients from a large population. In this paper, we propose a multimodal learning pipeline to predict PAWP marker. We utilize complementary information from Cardiac Magnetic Resonance Imaging (CMR) scans (short-axis and four-chamber) and Electronic Health Records (EHRs). We extract spatio-temporal features from CMR scans using tensor-based learning. We propose a graph attention network to select important EHR features for prediction, where we model subjects as graph nodes and feature relationships as graph edges using the attention mechanism. We design four feature fusion strategies: early, intermediate, late, and hybrid fusion. With a linear classifier and linear fusion strategies, our pipeline is interpretable. We validate our pipeline on a large dataset of $2,641$ subjects from our ASPIRE registry. The comparative study against state-of-the-art methods confirms the superiority of our pipeline. The decision curve analysis further validates that our pipeline can be applied to screen a large population. The code is available at \textcolor{blue}{\url{https://github.com/prasunc/hemodynamics}}.

\end{abstract}

\begin{IEEEkeywords}
Cardiac Hemodynamics, Feature Selection, Interpretable Model, Multimodal Learning, Pulmonary Arterial Wedge Pressure, Transformer. 
\end{IEEEkeywords}

\section{Introduction}
\label{S:Introduction}
Heart failure has emerged as a pressing global health concern, with its prevalence steadily rising over the past few years. The inability of the heart to supply adequate oxygen and blood to body organs is indicative of heart failure, which is a major cause of mortality and hospitalization~\citep{savarese2022global,emdin2009old}. Elevated Pulmonary Arterial Wedge Pressure (PAWP) indicates increased left ventricular filling pressure and impaired cardiac contractility. In the absence of pulmonary vasculature disease, PAWP correlates with the acuteness of heart failure and risk of hospitalization~\citep{adamson2014wireless}. In clinical practice, Right Heart Catheterization (RHC) is used as a reference method for the assessment of cardiovascular hemodynamics. However, expensive and invasive RHC procedure can not be used for screening patients at a population level. Therefore, simpler and non-invasive techniques can play a vital role in cardiovascular hemodynamics assessment~\citep {garg2022cardiac}.

 Cardiac Magnetic Resonance Imaging (CMR) is a promising tool for identifying various heart disorders~\cite{avendi2016combined}. A CMR scan provides a high spatio-temporal resolution that helps to effectively obtain images of the entire cardiac cycle. Machine Learning (ML) techniques were applied in several studies to perform automated diagnosis and prognosis from CMR images~\citep{assadi2022role,fotaki2022artificial}. In the literature, ML-based methods can be classified into two types. The first type of methods~\citep{alabed2022validation,vimalesvaran2022detecting} utilized clinical information or measurements extracted from CMR scans to perform prediction, whereas the second type of methods leveraged imaging features for the prediction. For instance, clinical features were used to predict Pulmonary Hypertension (PH) from CMR scans in~\citep{priya2021radiomics, alkhanfar2023non}. Over the years, several methods~\citep{diller2020utility,swift2021machine,uthoff2020geodesically,alabed2022machine,curiale2019automatic,gosling2023quantifying,puyol2020interpretable} were developed that work on imaging features extracted from CMR images. An ML-based method was developed in~\citep{swift2021machine} to diagnose Pulmonary Arterial Hypertension (PAH) using tensor-based features. Furthermore, mortality prediction was also performed using a similar type of pipeline in~\citep{uthoff2020geodesically,alabed2022machine}. With advancements in deep learning, several methods~\citep{curiale2019automatic,gosling2023quantifying,puyol2020interpretable, ahmadi2023transformer} utilized Convolutional Neural Network (CNN) features to perform the prediction. For example, CNN features were used to predict left ventricle function in~\citep{curiale2019automatic} and therapy response in~\citep{puyol2020interpretable}. In~\citep{ahmadi2023transformer}, researchers used a transformer-based method to predict aortic stenosis severity from cardiac data.

 \begin{figure*}[ht]
 \centering
\includegraphics[scale=0.57]{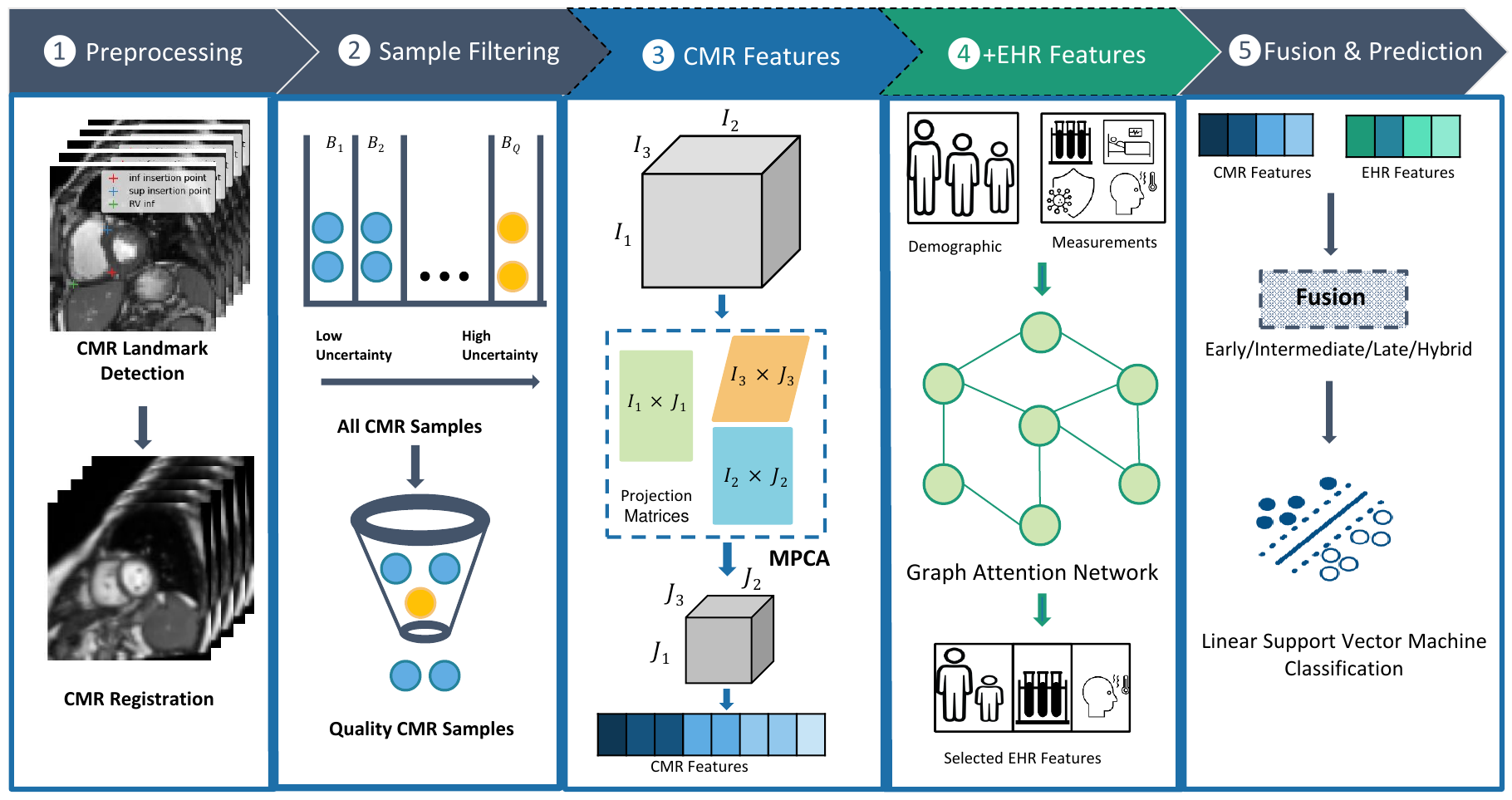}
\caption{The proposed multimodal pipeline for PAWP prediction utilizing features from short-axis CMR, four-chamber CMR, and EHR. Step 1: landmarks in CMRs are localized using Ensemble Maximum Heatmap Activation (E-MHA) strategy~\cite{schobs2022uncertainty} and CMRs are aligned to a common image space. Step 2: quality training samples are selected based on uncertainity scores. Step 3: spatio-temporal CMR features are extracted using MPCA~\citep{lu2008mpca}. Step 4: EHR features are selected using graph attention network. Step 5: the features are fused using early, intermediate, late, or hybrid fusion strategies. Then, the prediction is performed using linear Support Vector Machine (SVM).
}
\label{fig:model pipeline}
\end{figure*}

 CMR data contain spatio-temporal information collected for the entire cardiac cycle. The tensor-based methods~\citep{swift2021machine,uthoff2020geodesically,alabed2022machine} leverage Multilinear Principal Component Analysis (MPCA)~\citep{lu2008mpca} to extract spatio-temporal features from CMR scans. Most of the prior methods for CMRs deal with single-modality data. In recent years, multimodal learning has gained tremendous success in healthcare~\citep{soenksen2022integrated}. In this form of learning, complementary information from different modalities is leveraged to achieve enhanced performance. This motivated us to design a multimodal learning pipeline for PAWP prediction in our early conference paper~\citep{10.1007/978-3-031-43990-2_20} in which we fused the information from CMR images and Cardiac Measurements (CM)~\citep{garg2022cardiac}. In this paper, we substantially extend our pipeline by incorporating EHR features, enhancing feature fusion strategies, and validating on a large dataset.
The main contributions are summarized as follows: 

\begin{itemize}
    \item We developed an interpretable multimodal learning pipeline for PAWP prediction as depicted in Fig.~\ref{fig:model pipeline}. We utilized complementary information from features extracted from short-axis CMR, four-chamber CMR, and EHR data. We designed four feature fusion strategies that work on early, intermediate, late, and hybrid fusion. 
    \item We proposed a graph attention network (GAT) for EHR feature selection. We modeled subjects as graph nodes and feature relationships as edges.
    \item We performed uncertainty-based filtering to select quality samples for the training.
    
    \item We used a large dataset of $2,641$ subjects to perform experiments. We compared our pipeline with state-of-the-art methods~\citep{swift2021machine,garg2022cardiac,ahmadi2023transformer}. 
    
\end{itemize}
 
The remaining article is organized as follows. Materials and methods are presented in Section~\ref{ssec:Materials and Methods}. Experimental results are explained in Section~\ref{results}. The conclusion and future research are discussed in Section~\ref{conclusion}.

\section{Material and methods}
\label{ssec:Materials and Methods}
Figure~\ref{fig:model pipeline} depicts the pipeline for PAWP prediction. It consists of five steps including CMR preprocessing, sample filtering, CMR feature extraction, EHR feature selection, fusion and prediction. The sample filtering is used to eliminate low quality training samples based on uncertainty scores. We extract CMR features by utilizing Multilinear Principal Component Analysis (MPCA)~\citep{lu2008mpca}. We select important EHR features using a graph attention network and perform the prediction using Support Vector Machine (SVM). In this section, we discuss the dataset and each building block of the proposed pipeline.

\begin{table*}[!t]
\caption{Baseline characteristics of included patients. $p$ values were obtained using $t$-test~\cite{welch1947generalization}.\label{tab1}}
\centering
\setlength{\columnwidth}{0.1pt} 
\scalebox{1}{
\begin{tabular}{llll}
\toprule
& Low PAWP($\le15$) & High PAWP($>15$) & $p$-value\\
\midrule
Number of patients&$1612$&$1029$& -\\
\hline
Age (in years)&$63.2\pm 15.2$&$69.71\pm 11.2$& $<0.01$\\
\hline
Male (sex) &$612 (38\%)$&$452(44\%)$& $0.05$\\
\hline
Body Surface Area ($m^2$) &$1.91\pm 0.21$&$1.95\pm 0.22$& $<0.01$\\
\hline
Heart Rate (bpm) &$78.4\pm 14.1$&$70.3.6\pm 14.2$& $<0.01$\\
\hline
Left ventricular mass (m) &$91.3\pm 24$&$108\pm 31.1$& $<0.01$\\
\hline
Left atrial volume ($ml^{2}$) &$73.1\pm 31.7$&$131.1\pm 54.1$& $<0.01$\\
\hline
Left ventricular end-diastolic volume &$53.6\pm 14.9$&$62.1\pm 22.2$& $<0.01$\\
\hline
Left ventricular end-systolic volume &$19.4\pm 6.8$&$21.4\pm 11.3$& $<0.01$\\
\hline
Left ventricular stroke volume &$35.8 \pm 10.4$&$40.2\pm 14.4$& $<0.01$\\
\hline
Left ventricular ejection fraction ($\%$) &$49.2 \pm 13.4$&$55.3\pm 18.2$& $<0.01$\\
\hline
Right ventricular end-diastolic volume &$145.3\pm 58.8$&$153.1\pm 61.3$& $<0.01$\\
\hline
Right ventricular end-systolic volume &$81.5\pm 47.6$&$91.3\pm 44.7$& $<0.01$\\
\hline
Right ventricular stroke volume &$57.8 \pm 26.1$&$68.9\pm 27.7$& $<0.01$\\
\hline
Right ventricular ejection fraction ($\%$) &$42.4 \pm 12.6$&$44.4\pm 12.8$& $0.12$\\
\hline
Pulmonary Artery Wedge Pressure (mmHg) &$10.5\pm 3.4$&$21.5\pm 3.7$& $<0.01$\\
\bottomrule
\end{tabular}}
\end{table*}

\subsection{Dataset}
The dataset from Sheffield ASPIRE registry~\citep{hurdman2012aspire} was taken for this study. Patients with suspected pulmonary hypertension were identified after institutional review board approval and ethics committee review. This dataset was collected from the year $2011$ to $2023$. A total of $2,641$ patients who underwent Right Heart Catheterization (RHC) and CMR scans within $24$ hours were selected. Among these patients, $1,612$ exhibited normal PAWP ($\leq15$ mmHg), while $1,029$ had elevated PAWP ($>15$ mmHg). RHC was conducted using a balloon-tipped $7.5$ French thermodilution catheter. PAWP was estimated using standard techniques and averaged over multiple cardiac cycles to avoid any overestimation.  Table~\ref{tab1} shows the baseline characteristics of subjects.
 
 CMR scans were obtained using a $1.5$ Tesla whole-body GE HDx MRI scanner (GE Healthcare, Milwaukee, USA) equipped with retrospective electrocardiogram gating and $8$-channel cardiac coils. Two CMR sequences were utilized: the short-axis and four-chamber. These sequences adhered to standard clinical guidelines and involved the acquisition of cardiac-gated multi-slice steady-state sequences. The acquired images had a slice thickness of $8$ mm, a field of view measuring $48$ mm by $43.2$ mm, a matrix size of $512\times 512$, a bandwidth of $125$ kHz, and a repetition time and echo time of $3.7$ ms and $1.6$ ms, respectively.

\subsection{Preprocessing}
The preprocessing of CMR images contains two steps: automated landmark detection and registration. We leverage a CNN-based method to detect landmarks~\cite{schobs2022uncertainty} from CMR scans.
We use the U-Net-like CNN and leverage the same training strategy implemented in~\cite{schobs2022uncertainty}. We utilize \textit{Ensemble Maximum Heatmap Activation (E-MHA)} strategy~\cite{schobs2022uncertainty} to locate the landmarks. This strategy incorporates an ensemble of five models for each modality. We leverage three landmarks for each modality. For short-axis modality, we use the inferior hinge point, superior hinge point, and inferolateral inflection point of the right ventricular apex. For the four-chamber modality, we utilize the left ventricular apex and mitral and tricuspid annulus. E-MHA strategy also produces an associated uncertainty score for each landmark that represents the epistemic uncertainty of the prediction in the form of a continuous scalar value. We carry out affine registration on the scans after predicting the landmarks. The image registration helps to align the heart regions of different subjects to a common image space.

\subsection{Training Sample Selection}
We aim to use quality samples for the training of our model. A minor deviation in landmark prediction can lead to incorrect image registration.
We hypothesize that incorrectly registered cardiac scans resulting from inaccurate landmarks can impact the training process, resulting in a sub-optimal model. Therefore, it becomes crucial to identify and handle such samples for quality control. We design a simple and effective strategy to discard low-quality samples. We leverage predicted landmarks with epistemic uncertainty scores to tackle this problem as in~\cite{schobs2022uncertainty}. We discard the training samples based on the uncertainty scores of the landmarks. The predicted landmarks are partitioned into $Q$ quantiles, i.e., $\text{B}=\{B_{1},B_{2},...,B_{Q}\}$, based on the epistemic uncertainty scores. The training samples are then iteratively filtered out starting from the most uncertain partition (i.e., $B_{Q}$). A training sample is removed if the uncertainty score of any of its landmarks falls in the remaining most uncertain quantile at iteration $\rho$. The process of discarding samples is carried out iteratively until there is no observable enhancement in the validation performance, as quantified by the Area Under Receiver Operating Characteristic Curve (AUROC), over two consecutive iterations.


\subsection{CMR Feature Extraction}
CMR scans contain spatio-temporal information for the whole cardiac cycle. Multilinear Principal Component Analysis (MPCA)~\citep{lu2008mpca} is a well-established method to extract tensor-based features from CMR data~\citep{swift2021machine,uthoff2020geodesically,alabed2022machine}. The MPCA method learns multilinear bases from CMR stacks to obtain low-dimensional tensors that contain spatio-temporal features for the prediction. Suppose we have a set of $M$ scans represented as third-order tensors in the form of $\{\mathcal{X}_1,\mathcal{X}_2,..,\mathcal{X}_M\in \mathbb{R}^{I_{1} \times I_{2} \times I_{3}}\}$. The low-dimensional tensor features $\{\mathcal{Y}_1,\mathcal{Y}_2,..,\mathcal{Y}_M\in \mathbb{R}^{J_{1} \times J_{2} \times J_{3}}\}$ are obtained by learning three ($N=3$) projection matrices $\{\mathbf{U}^{(n)}\in \mathbb{R}^{I_{n} \times J_{n} }, n=1,2,3\}$ as follows:

\begin{equation}
    \mathcal{Y}_m=\mathcal{X}_{m} \times_{1} \mathbf{U}^{(1)^\top} \times_{2} \mathbf{U}^{(2)^\top}\times_{3} \mathbf{U}^{(3)^\top},m=1,2,...,M,
\end{equation}
where $J_{n}<I_{n}$, and $\times_n$ denotes a mode-wise product. In this way, the feature dimensions are reduced from $I_{1} \times I_{2} \times I_{3}$ to $J_{1} \times J_{2} \times J_{3}$. We optimize the projection matrices $\{\mathbf{U}^{(n)}\}$ by maximizing the total scatter $\psi_{\mathcal{Y}}=\sum_{m=1}^M||\mathcal{Y}_{m}-\bar{\mathcal{Y}}||_{F}^{2}$, where $\bar{\mathcal{Y}}=\frac{1}{M}\sum_{m=1}^M\mathcal{Y}_{m}$ is the mean tensor feature and $||.||_{F}$ is the Frobenius norm~\cite{lu2013multilinear}. This problem is solved via an iterative projection method. In MPCA, the explained variance ratio allows us to determine $\{J_{1},J_{2},J_{3}\}$. We further rank the obtained features using Fisher discriminant analysis~\cite{li2017feature}.  We keep only the top $\kappa$-ranked features for the prediction.

\begin{figure*}[t]
\centering
\includegraphics[scale=0.81, trim=3cm 0.5cm 4cm 0cm]{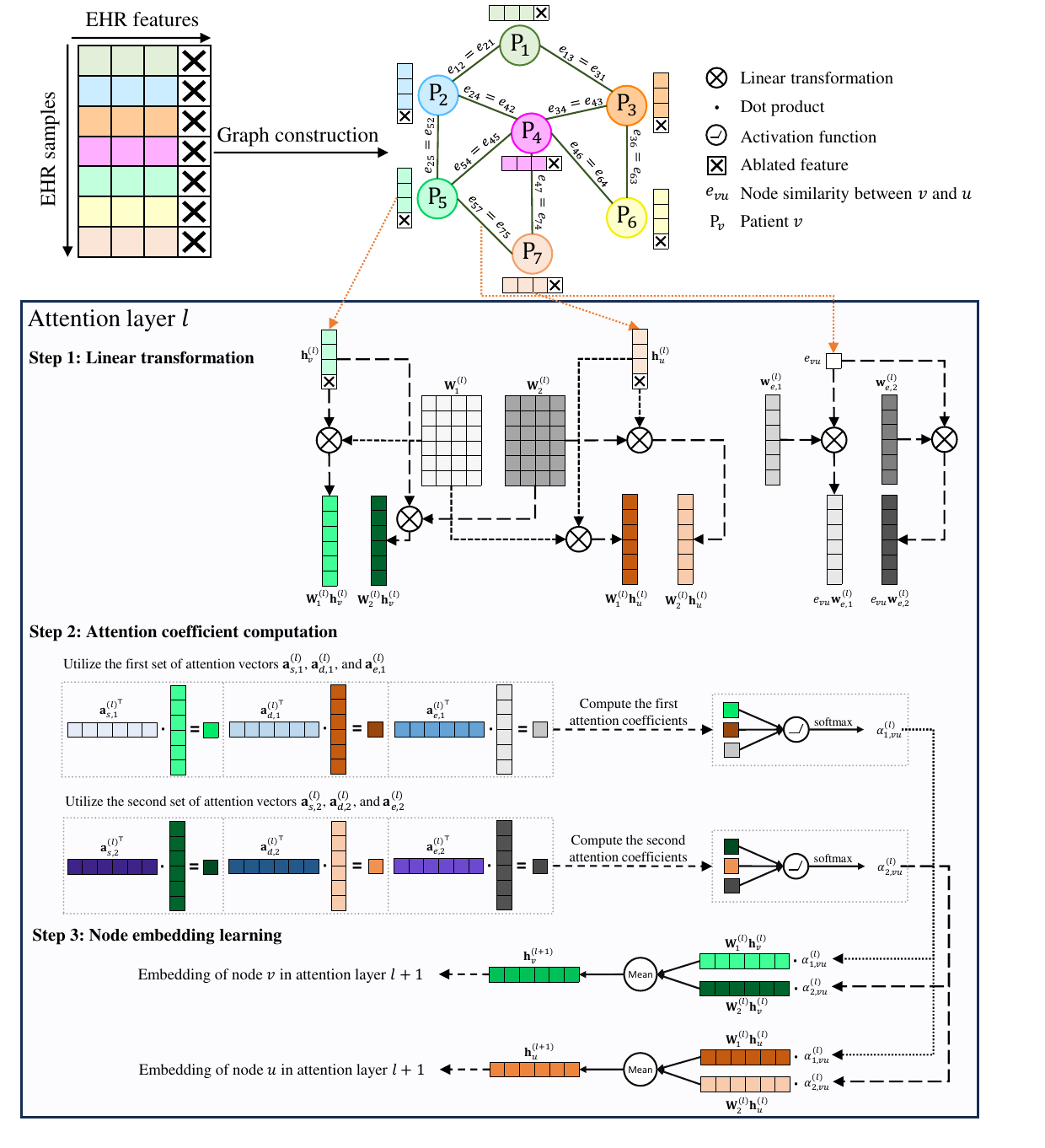}
\caption{EHR feature selection based on GAT and the ablation approach ~\citep{wang2021mogonet}. \textbf{Top:} Input EHR data and the corresponding constructed graph. \textbf{Bottom:} A single attention layer in GAT with two attention heads applied to EHR nodes in three steps. In step 1, nodes are linearly transformed for high-level feature embedding. In step 2, the attention mechanism utilizes attention vectors to compute normalized attention coefficients of the attention matrix $\mathbf{A}^{\left(l\right)}$ on these embeddings. In step 3, these coefficients are used to perform a linear combination of node embeddings, resulting in the final node embeddings for the next attention layer.}
\label{fig:gat_model}
\end{figure*}

\begin{figure}[!t]
	\centering
	\begin{subfigure}[b]{0.48\textwidth}
		\centering
		\includegraphics[width=\textwidth]{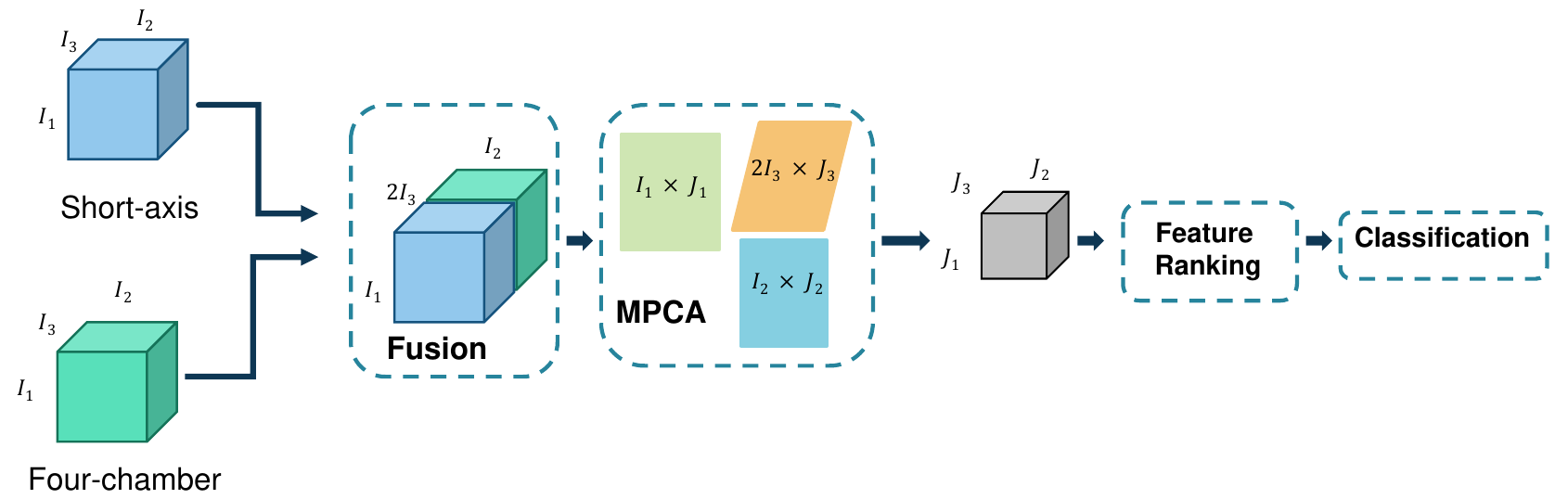}
		\caption{Early fusion.\label{fig2a}}
	\end{subfigure}
 \vspace{0cm}
 \begin{subfigure}[b]{0.48\textwidth}
		\centering
		\includegraphics[width=\textwidth]{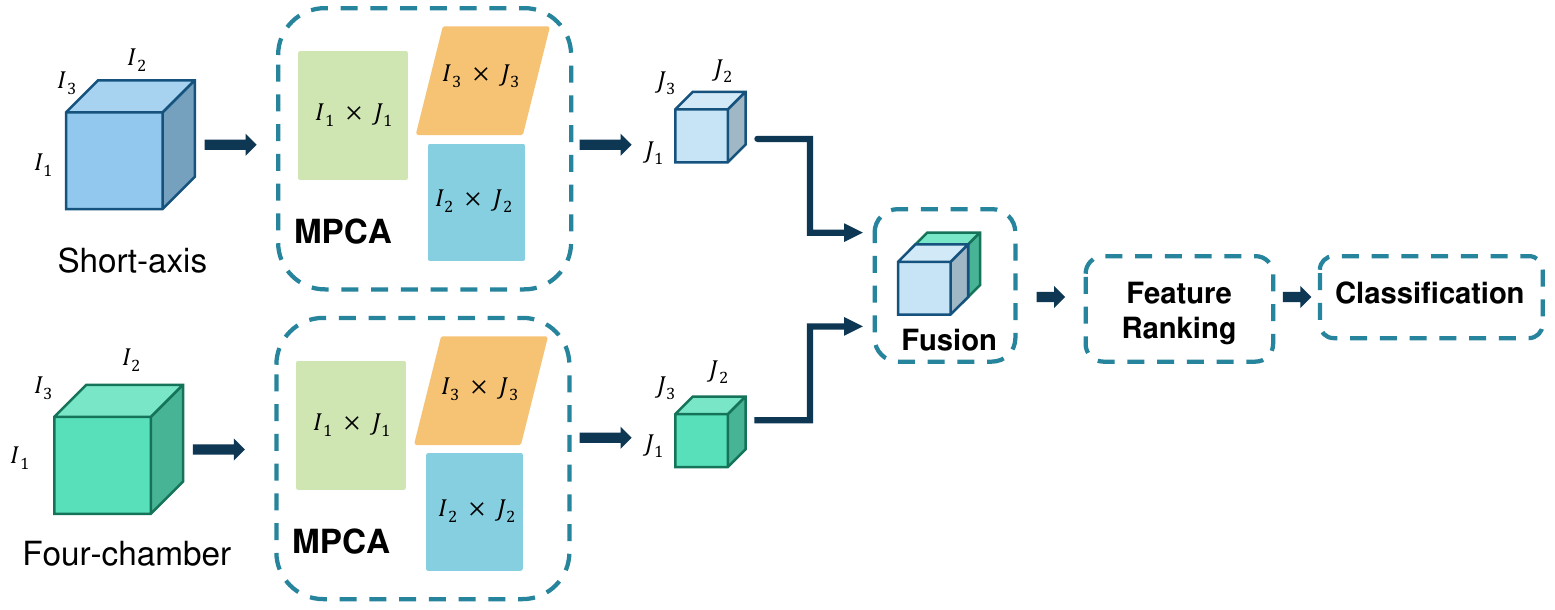}
		\caption{Intermediate fusion.\label{fig2b}}
	\end{subfigure}
 \vspace{0cm}
 \begin{subfigure}[b]{0.48\textwidth}
		\centering
		\includegraphics[width=\textwidth]{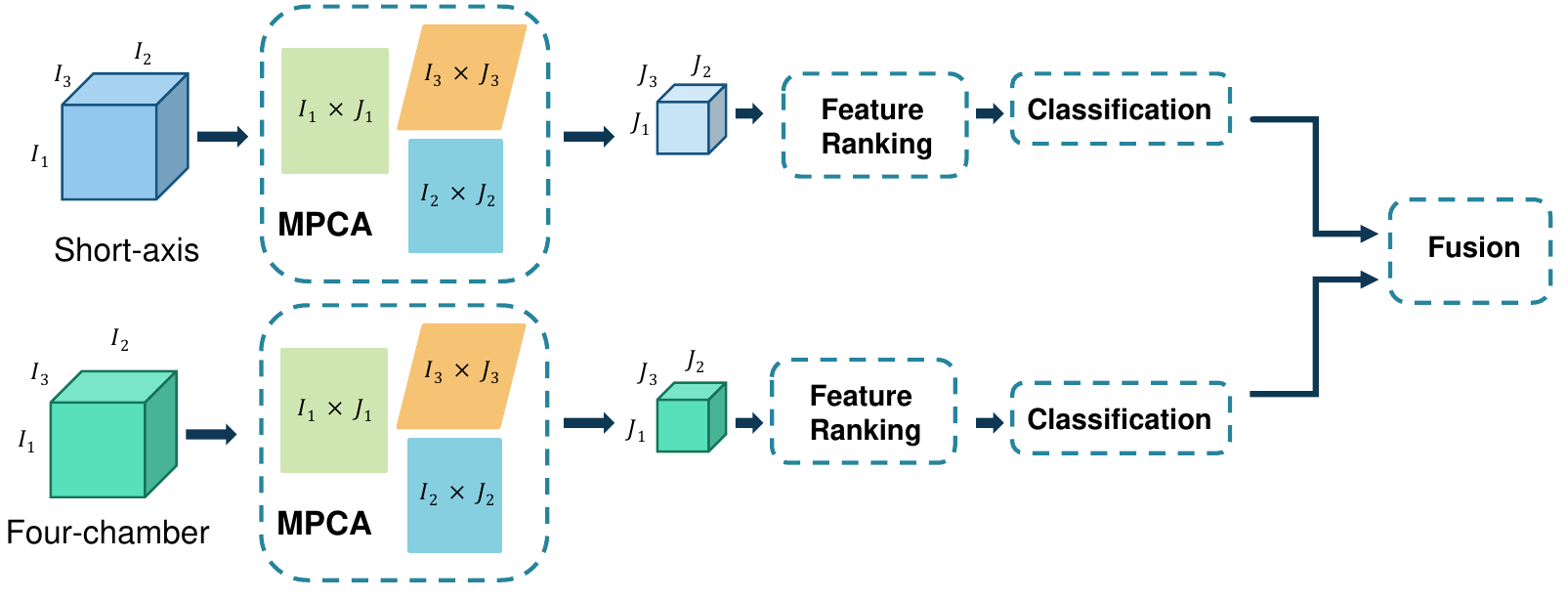}
		\caption{Late fusion.\label{fig2c}}
	\end{subfigure}

 \begin{subfigure}[b]{0.48\textwidth}
		\centering
		\includegraphics[width=\textwidth]{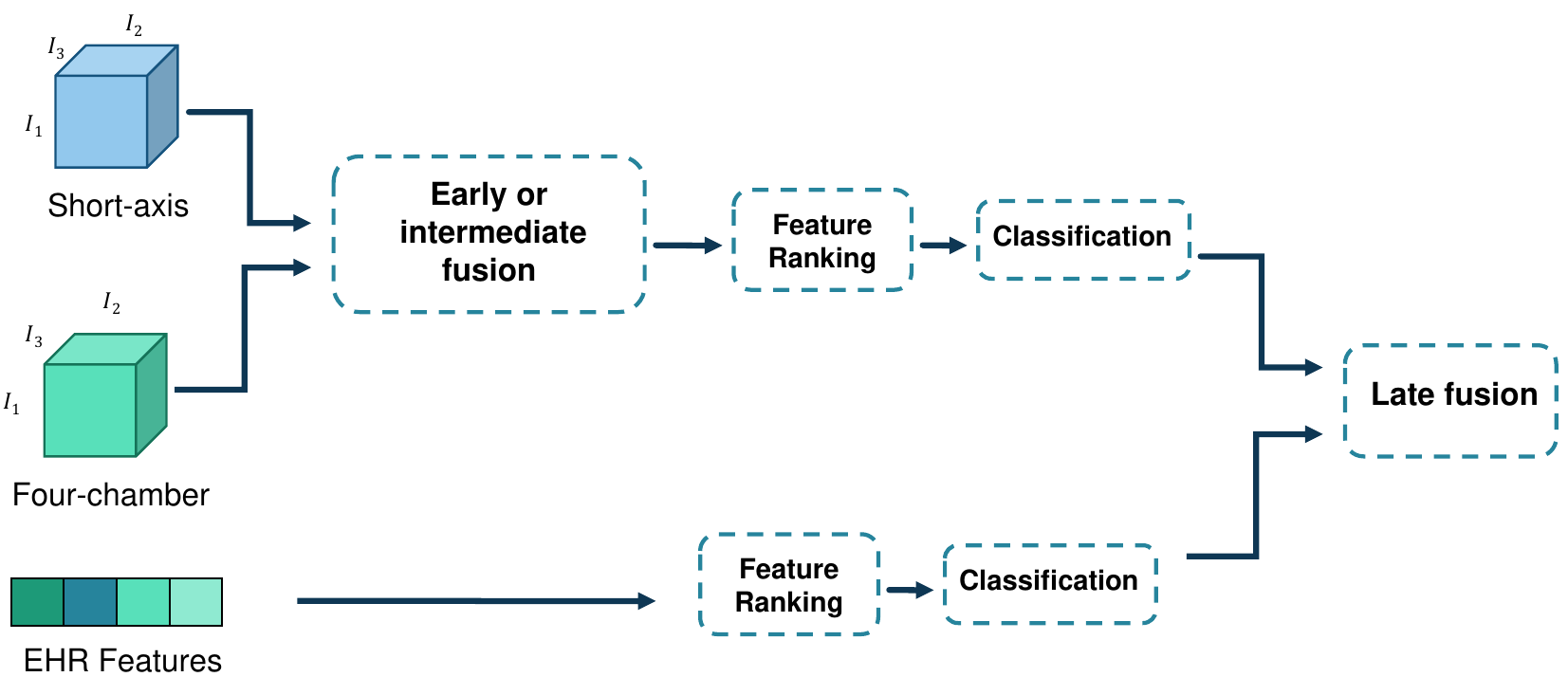}
		\caption{Hybrid fusion.\label{fig2d}}
	\end{subfigure}
\caption{Four types of fusion methods utilized in our pipeline. }\label{fig2}
\end{figure}

\subsection{EHR Feature Selection with Graph Attention Network}
To automate the extraction of representative features from EHR data, we propose a feature selection approach leveraging the power of graph attention network (GAT)~\citep{veličković2018graph}, as shown in Fig.~\ref{fig:gat_model}. GAT extends deep neural networks to graph-structured data for effectively capturing complex relationships within data. We construct a graph using EHR data where nodes represent patients and edges represent their relationships. This enables us to apply GAT to the constructed graph to identify salient EHR features based on their relative importance in the overall network. The fundamental idea behind attention mechanisms lies in the ability to compute node representations by assigning different importance or attention weights to the neighborhoods connected to the target node~\citep{veličković2018graph, hamilton2020graph}. The construction of GAT involves stacking multiple attention layers, with each layer defined as follows:

\begin{equation}
\mathbf{H}^{\left(l+1\right)}=\mathbf{A}^{\left(l\right)} \mathbf{H}^{\left(l\right)}{\mathbf{W}^{\left(l\right)}}^\top,
\end{equation}
where $\mathbf{A}^{\left(l\right)}\in\mathbb{R}^{V\times V}$ is the attention mechanism matrix at the layer $l$, $V$ is the number of nodes in the graph, $\mathbf{H}^{\left(l\right)}\in\mathbb{R}^{V\times d^{(l)}}$ is the node representations at the layer $l$, $\mathbf{W}^{\left(l\right)}\in\mathbb{R}^{d^{(l+1)}\times d^{(l)}}$ is a trainable weight matrix designed to transform node representations into another embedding space. Here, $d^{(l)}$ and $d^{(l+1)}$ are the hidden-layer dimensions in layer $l$ and layer $l$+1, respectively, and $(.)^\top$ represents transposition. The initial node representations correspond to the input node feature matrix, denoted as $\mathbf{H}^{\left(0\right)}=\mathbf{D}$, where  $\mathbf{D}\in\mathbb{R}^{V\times d^{(0)}}$, and $d^{(0)}$ indicates the input feature dimensionality. The attention mechanism computes each coefficient $\alpha_{vu}^{(l)}$ of the attention matrix $\mathbf{A}^{\left(l\right)}$ according to the following expression:

\begin{equation}
\!\!\!\!\alpha_{vu}^{(l)}\!=\!\frac{
    \exp\!{\left(\sigma\!\left(\substack{
    {\mathbf{a}}_{s}^{{(l)}^\top}\!\mathbf{W}^{(l)} \mathbf{h}_v^{(l)} \!+
    {\mathbf{a}}_{d}^{{(l)}^\top}\!\mathbf{W}^{(l)} \mathbf{h}_u^{(l)} \!+
    {e_{vk} \mathbf{a}}_{e}^{{(l)}^\top}\!\mathbf{w}_{e}^{(l)}}\right)\right)}}
    {\!\!\!\sum\limits_{k\in \mathcal{N}_v}\!\!
    \exp\!{\left(\sigma\!\left(\substack{
    {\mathbf{a}}_{s}^{{(l)}^\top}\!\mathbf{W}^{(l)} \mathbf{h}_v^{(l)} \!+
    {\mathbf{a}}_{d}^{{(l)}^\top}\!\mathbf{W}^{(l)} \mathbf{h}_k^{(l)} \!+
    {e_{vk} \mathbf{a}}_{e}^{{(l)}^\top}\!\mathbf{w}_{e}^{(l)}}\right)\right)}},\!\!
\end{equation}
where $\mathbf{h}_v^{(l)}\in\mathbb{R}^{d^{(l)}}$ and $\mathbf{h}_u^{(l)}\in\mathbb{R}^{d^{(l)}}$ denote the hidden representations of nodes $v$ and $u$, respectively, derived from node representation matrix $\mathbf{H}^{\left(l\right)}$, $e_{vu}$ indicates the cosine similarity between nodes $v$ and $u$ in the adjacency matrix, $\mathcal{N}_v$ denotes the set of directly connected neighbors of node $v$ in the graph, as defined by the adjacency matrix, ${\mathbf{a}}_{s}^{(l)}$, ${\mathbf{a}}_{d}^{(l)}$, and ${\mathbf{a}}_{e}^{(l)}$ are learnable attention vectors for the source, destination, and edge features, respectively, $\sigma(.)$ is the LeakyReLU nonlinearity function, and $\exp(.)$ is the standard exponential function. The adjacency matrix captures the structure of the sample similarity network by calculating the cosine similarity between node pairs. As suggested by Wang et al.~\citep{wang2021mogonet}, we preserve only those edges where the cosine similarity is larger than a specified threshold $\theta_e$. This threshold corresponds to the average number of retained edges per node. 

To enhance the attention mechanism's learning, the computation of attention weights involves multiple heads, an approach known as multi-head attention~\citep{veličković2018graph}. This is achieved by utilizing independent attention mechanisms, resulting in the following node representations:

\begin{equation}
\mathbf{H}^{\left(l+1\right)}=
    \frac{1}{K}\sum_{k=1}^{K}{\mathbf{A}_k^{\left(l\right)} \mathbf{H}^{\left(l\right)}{\mathbf{W}_k^{\left(l\right)}}^\top},
\end{equation}
where $K$ is the number of attention heads.

The GAT model encodes the input matrix $\mathbf{D}$ and its adjacency matrix into low-dimensional node embeddings through a set of attention layers. These node representations are subsequently fed into a single-layer neural network acting as a decoder to produce the final output. 

To determine the significance of each feature, we employ the ablation approach, a commonly used technique in neural networks~\citep{wang2021mogonet, setiono1997neural, 9468895}. Following the training of our GAT model, we individually eliminate each feature from the input data by setting its values to zero and observe the performance difference (i.e., $\Delta$AUROC) on the validation set. The significant decrease in performance indicates the importance of the feature. We rank the features based on their importance and select the top $\Theta$ ones for our pipeline.




\subsection{Multimodal Feature Fusion}
We leverage the complementary information from the features of the four-chamber CMR, short-axis CMR, and EHR data. In EHR data, we included $5$ demographic and $53$ clinical measurement features. We eliminated $9$ features which contain more than $5\%$ missing values. We imputed the missing values of the remaining features with the mean value. In total, we kept $49$ features from the EHR data. We applied GAT-based feature selection to select $15$ most important EHR features.

We fuse features using four strategies: early, intermediate, late, and hybrid fusion. Fig.~\ref{fig2a} depicts the early fusion of two imaging modalities. Suppose we have $M$ scans of short-axis and four-chamber modalities represented as $\{\mathcal{X}_{1}^{S},\mathcal{X}_{2}^{S},..,\mathcal{X}_{M}^{S}\in \mathbb{R}^{I_{1} \times I_{2} \times I_{3}}\}$ and $\{\mathcal{X}_{1}^{F},\mathcal{X}_{2}^{F},..,\mathcal{X}_{M}^{F}\in \mathbb{R}^{I_{1} \times I_{2} \times I_{3}}\}$, respectively. In early fusion, we concatenate these tensors before applying MPCA. Therefore, we obtain tensors $\{\mathcal{X}_{1}^{SF},\mathcal{X}_{2}^{SF},..,\mathcal{X}_{M}^{SF}\in \mathbb{R}^{I_{1} \times I_{2} \times 2I_{3}}\}$. We apply MPCA on the concatenated tensors and obtain $\{\mathcal{Y}_1,\mathcal{Y}_2,..,\mathcal{Y}_M\in \mathbb{R}^{J_{1} \times J_{2} \times J_{3}}\}$. We then rank and select features using the Fisher score and perform the prediction.


In intermediate fusion, we apply MPCA on both imaging modalities and then combine tensors at the latent space as shown in Fig.~\ref{fig2b}. We obtain low-dimensional tensors $\{\mathcal{Y}_{1}^{S},\mathcal{Y}_{2}^{S},..,\mathcal{Y}_{M}^{S}\in \mathbb{R}^{J_{1} \times J_{2} \times J_{3}}\}$ and $\{\mathcal{Y}_{1}^{F},\mathcal{Y}_{2}^{F},..,\mathcal{Y}_{M}^{F}\in \mathbb{R}^{J_{1} \times J_{2} \times J_{3}}\}$ for short-axis and four-chamber modalities. To this end, we combine these tensors to obtain tensors $\{\mathcal{Y}_{1}^{SF},\mathcal{Y}_{2}^{SF},..,\mathcal{Y}_{M}^{SF}\in \mathbb{R}^{J_{1} \times J_{2} \times 2J_{3}}\}$. We then select top-ranked features for the prediction.

 As shown in Fig.~\ref{fig2c}, the late fusion is straightforward in which we fuse decision scores obtained from individual models trained for separate modalities. The above early and intermediate fusion can only be applied to modalities that have the same dimensionalities, whereas late fusion can also be applied to modalities that have different dimensionalities. Therefore, early and intermediate fusion can be used to combine short-axis and four-chamber modalities only but late fusion can be utilized to combine short-axis, four-chamber, and EHR modalities.
 
 We perform hybrid fusion to combine three modalities as depicted in Fig.~\ref{fig2d}. In this fusion, we use one type of fusion (i.e., early or intermediate) to combine the first and second modalities, whereas we apply a second type of fusion (i.e., late) to integrate the third modality. As shown in Fig.~\ref{fig2d}, we first perform early or intermediate fusion to combine short-axis and four-chamber modalities, and then carry out late fusion with EHR features. Our fusion strategies work on linear combinations of features, enabling us to make our pipeline fully interpretable. After feature fusion, we perform the prediction using a Support Vector Machine (SVM) with a linear kernel.

\section{Results and Analysis}
\label{results}
This section presents the experimental design, results analysis, ablation studies, and feature interpretation.
\subsection{Experimental Design}
We performed experiments on three modalities including short-axis CMR, four-chamber CMR, and EHR features. We used grid search to select optimal hyperparameters in linear SVM. We chose hyperparameters from \{$0.001, 0.01, 0.1, 1$\}. We used $10$-fold cross-validation in the training set to select optimal hyperparameters. We selected $210$ tensor features from CMR scans and $15$ features from EHR data. To select relevant EHR features, we employed a three-layer GAT model with hidden dimensions of $[64, 64, 64]$, each equipped with three attention heads. Furthermore, we utilized LeakyReLU nonlinearity with a negative input slope of $0.25$ followed by dropout~\citep{JMLR:v15:srivastava14a} with probability $p$ = $0.5$ for all layers. \textit{The use of the LeakyReLU nonlinearity does not compromise the interpretability, as the selected EHR features can be directly incorporated in our complete pipeline}. The GAT model was implemented in PyTorch Geometric~\citep{FeyLenssen2019} with the Adam optimizer~\citep{kingma2015adam} and a learning rate of $0.01$, and trained with cross-entropy loss over $400$ epochs. During adjacency matrix construction, we retained an average of $10$ edges per node~\citep{wang2021mogonet}.

We selected $2,641$ subjects who had all three modalities and partitioned them into a training set of $1,849$ and a test set of $792$. To simulate a real testing scenario, patients screened in early years were taken in the training set, while those diagnosed in recent years were included in the test set. Moreover, we partitioned the test set into five segments according to screening time, enabling us to execute multiple studies of methods and present standard deviations in the comparison results. In this work, we utilized three performance metrics: Area
Under Receiver Operating Characteristic Curve (AUROC), accuracy, and Matthew’s Correlation Coefficient (MCC)~\citep{chicco2020advantages}. We evaluated the performance of our pipeline on three CMR resolutions: $128\times 128$, $64\times 64$, and $32\times 32$. All experiments were performed in Python (version $3.9$). MPCA and SVM implementations were adopted from Python libraries PyKale~\citep{lu2022pykale} and scikit-learn~\citep{pedregosa2011scikit}, respectively. Baseline EHR feature selection methods included Fisher score~\citep{10.5555/954544}, minimal-redundancy–maximal-relevance (mRMR)~\citep{mrmr2025peng} from the scikit-feature package~\citep{li2018feature}, and recursive feature elimination (RFE)~\citep{guyon2002gene} from scikit-learn. 

Our pipeline was compared with two clinical baselines~\citep{swift2021machine,garg2022cardiac} and one deep learning-based method~\citep{ahmadi2023transformer}. To compare our work with deep learning, we performed the comparison with a transformer-based method~\citep{ahmadi2023transformer}. This method was developed to predict aortic stenosis from spatio-temporal cardiac data. We used ResNet18 to obtain spatial features. The input resolution was fixed to $224\times 224$. We trained the transformer method for $100$ epochs with Adam optimizer~\citep{kingma2015adam} and $0.001$ learning rate. For fair comparison, we used the same preprocessing of cardiac data (i.e., landmark detection and uncertainty-based filtering) for all comparing methods.

\begin{figure}[!t]
\centering
\includegraphics[scale=0.38]{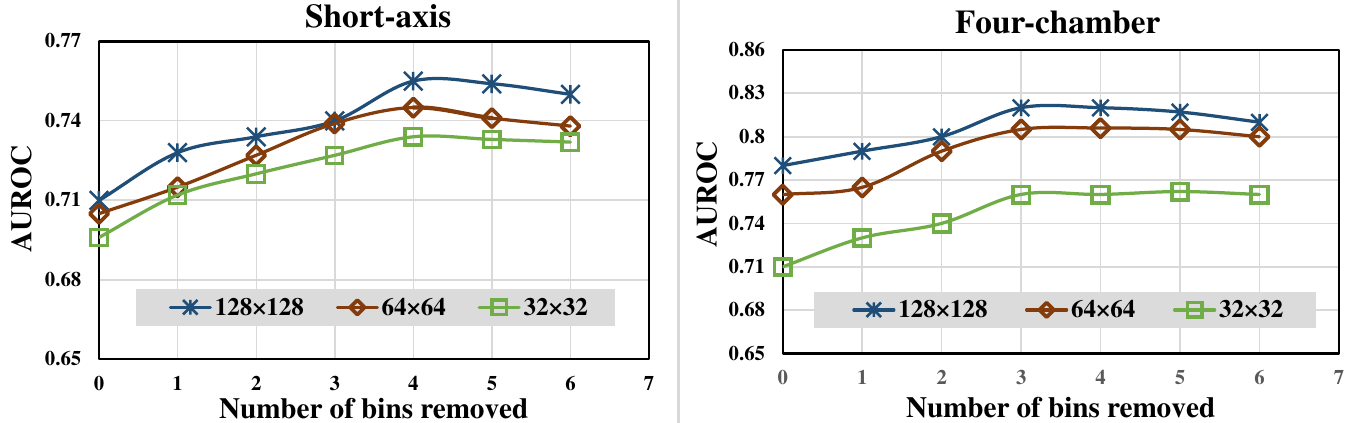}
\caption{The validation performance on removing different numbers of bins from training data.}
\label{fig3}
\end{figure}

\begin{table*}[!t]
\caption{Performance comparison of unimodal models using three metrics (with \textcolor{red}{best} in Red and \textcolor{blue}{second best} in Blue). FC: Four-Chamber features; SA: Short-Axis features; CM: Cardiac Measurement features; EHR: Electronic Health Records.   }\label{tab2}
\centering
\scalebox{0.97}{
\begin{tabular}{l|r|c|c|c}
\hline
Method (Modality) & Resolution & AUROC & Accuracy & MCC\\
\hline
Baseline (CM)~\cite{garg2022cardiac} &-&$0.7300\pm 0.04$&$0.7400\pm 0.03$&$0.2081 \pm 0.03$\\
\hline
Baseline (EHR) &-&$0.7741\pm 0.05$&$0.7905\pm 0.06$&$0.3251 \pm 0.03$\\
\hline
       MPCA (SA)~\citep{swift2021machine} &$32\times32$&$0.7219\pm 0.05$&$0.7218 \pm 0.06$&$0.2783\pm0.03$ \\
       &$64\times64$&$0.7357\pm 0.06$&$0.7413\pm 0.05$&$0.3074\pm0.02$ \\
        &$128\times128$&$0.7391\pm 0.04$&$0.7475\pm 0.03$&$0.3211\pm0.03$ \\
        \hline
 MPCA (FC)~\citep{swift2021machine}    &$32\times32$&$0.7774 \pm 0.04$&$0.7548 \pm 0.07$&$0.3224\pm0.04$ \\
&$64\times64$&$0.8042 \pm 0.06$&$0.7723 \pm 0.04$&$0.4217\pm0.04$ \\
     &$128\times128$&\textcolor{blue}{$0.8095 \pm 0.04$}&$\textcolor{red}{0.7912 \pm 0.03}$&\textcolor{blue}{$0.4521\pm0.04$} \\
        \hline
  Transformer (SA)~\citep{ahmadi2023transformer}    &$224\times224 $&$0.7416\pm0.02$&$0.7492\pm0.06$&$0.3196\pm0.04$ \\
  \hline
 Transformer (FC)~\citep{ahmadi2023transformer}   &$224\times224$&$\textcolor{red}{0.8124\pm0.05}$&\textcolor{blue}{$0.7831\pm0.06$}&$\textcolor{red}{0.4584\pm0.04}$ \\
 \hline
\end{tabular}}
\end{table*}

 \begin{table*}[!t]
\caption{Performance comparison of bi-modal models using three metrics.   }\label{tab3}
\centering
\scalebox{0.97}{
\begin{tabular}{l|l|r|c|c|c}
\hline
Method (Modalities) &Fusion& Resolution & AUROC & Accuracy & MCC\\
\hline
 MPCA (SA $\&$ FC) &Early&$32\times32$&$0.7351\pm 0.06$&$0.7416 \pm 0.03$&$0.3038\pm0.04$ \\
&&$64\times64$&$0.7491\pm 0.07$&$0.7510\pm 0.06$&$0.3417\pm0.04$ \\
&&$128\times128$&$0.7490\pm 0.04$&$0.7527\pm 0.04$&$0.3586\pm0.03$ \\
        \hline
MPCA (SA $\&$ FC) &Intermediate&$32\times32$&$0.7816\pm 0.07$&$0.7452 \pm 0.08$&$0.3681\pm0.04$ \\
&&$64\times64$&$0.8115\pm 0.07$&$0.7671\pm 0.04$&$0.4280\pm0.03$ \\
&&$128\times128$&$\textcolor{red}{0.8269\pm 0.05}$&$\textcolor{red}{0.7998\pm 0.04}$&$\textcolor{red}{0.4748\pm0.04}$ \\
        \hline
MPCA (SA $\&$ FC) &Late&$32\times32$&$0.7802\pm 0.06$&$0.7568 \pm 0.05$&$0.3418\pm0.02$ \\
&&$64\times64$&$0.8078\pm 0.04$&$0.7614\pm 0.07$&$0.4168\pm0.04$ \\
&&$128\times128$&$0.8129\pm 0.05$&$0.7881\pm 0.05$&$0.4681\pm0.02$ \\
        \hline

MPCA (SA $\&$ EHR) &Late&$32\times32$&$0.7435\pm 0.05$&$0.7581 \pm 0.04$&$0.2981\pm0.04$ \\
&&$64\times64$&$0.7615\pm 0.04$&$0.7795\pm 0.05$&$0.3352\pm0.03$ \\
&&$128\times128$&$0.7809\pm 0.05$&$0.7825\pm 0.04$&$0.3478\pm0.05$ \\
        \hline
MPCA (FC $\&$ EHR) &Late&$32\times32$&$0.7852\pm 0.05$&$0.7641 \pm 0.05$&$0.3648\pm0.03$ \\
&&$64\times64$&$0.8082\pm 0.06$&$0.7951\pm 0.06$&$0.4487\pm0.03$ \\
&&$128\times128$&$0.8161\pm 0.05$&$0.7968\pm 0.05$&$0.4628\pm0.02$ \\
        \hline
Transformer (SA $\&$ FC)~\citep{ahmadi2023transformer} &Early&$224\times224 $&$0.8158\pm 0.07$&$0.7892 \pm 0.05$&$0.4595\pm0.04$ \\
\hline
Transformer (SA $\&$ FC)~\citep{ahmadi2023transformer} &Late&$224\times224 $&$0.8241\pm 0.04$&$0.7910 \pm 0.05$&$0.4617\pm0.04$ \\
\hline
Transformer (SA $\&$ EHR)~\citep{ahmadi2023transformer} &Late&$224\times224 $&$0.7780\pm 0.05$&$0.7852 \pm 0.06$&$0.3381\pm0.03$ \\
\hline
 Transformer (FC $\&$ EHR)~\citep{ahmadi2023transformer} &Late&$224\times224 $&$\textcolor{blue}{0.8255\pm 0.05}$&$\textcolor{blue}{0.7971 \pm 0.05}$&$\textcolor{blue}{0.4650\pm0.02}$ \\
 \hline
\end{tabular}}
\end{table*}

\begin{table*}[!t]
\caption{Performance comparison of tri-modal models using three metrics.}\label{tab4}
\centering
\scalebox{.97}{
\begin{tabular}{l|l|r|c|c|c}
\hline
Method (Modalities) &Fusion& Resolution & AUROC & Accuracy & MCC\\
\hline
MPCA (SA, FC, $\&$ EHR) &Hybrid: Early (SA, FC) $\&$ Late (EHR)&$32\times32$&$0.8050\pm 0.06$&$0.7852 \pm 0.06$&$0.4892\pm0.04$ \\
&&$64\times64$&$0.8271\pm 0.07$&$0.8156\pm 0.05$&$0.5075\pm0.03$ \\
&&$128\times128$&$0.8610\pm 0.05$&$0.8405\pm 0.04$&$0.5286\pm0.02$ \\
        \hline
MPCA (SA, FC, $\&$ EHR) & Hybrid: Intermediate (SA, FC) $\&$ Late 
 (EHR)&$32\times32$&$0.8162\pm 0.04$&$0.8098 \pm 0.05$&$0.5012\pm0.03$ \\
&&$64\times64$&$0.8508\pm 0.05$&$0.8302\pm 0.06$&$0.5251\pm0.04$ \\
&&$128\times128$&$\textcolor{red}{0.8682\pm 0.04}$&$\textcolor{red}{0.8601\pm 0.03}$&$\textcolor{red}{0.5492\pm0.03}$ \\
        \hline
MPCA (SA, FC, $\&$ EHR) &Late (SA, FC, $\&$ EHR)&$32\times32$&$0.8246\pm 0.05$&$0.8341 \pm 0.04$&$0.4898\pm0.04$ \\
&&$64\times64$&$0.8402\pm 0.04$&$0.8327\pm 0.05$&$0.5245\pm0.03$ \\
&&$128\times128$&\textcolor{blue}{$0.8625\pm 0.05$}&\textcolor{blue}{$0.8422\pm 0.06$}&\textcolor{blue}{$0.5381\pm0.04$} \\
        \hline
Transformer (SA, FC, $\&$ EHR)~\citep{ahmadi2023transformer} &Hybrid: Early (SA, FC) $\&$ Late (EHR)&$224\times224 $&$0.8582\pm 0.05$&$0.8376 \pm 0.05$&$0.5348\pm0.03$ \\
\hline
Transformer (SA, FC, $\&$ EHR)~\citep{ahmadi2023transformer} &Late (SA, FC, $\&$ EHR)&$224\times224 $&$0.8498\pm 0.06$&$0.8297 \pm 0.06$&$0.5217\pm0.02$ \\
 \hline
\end{tabular}}
\end{table*}

\begin{figure*}[t]
	\centering
	\begin{subfigure}[b]{0.48\textwidth}
		\centering
		\includegraphics[width=\textwidth, height=2.75cm]{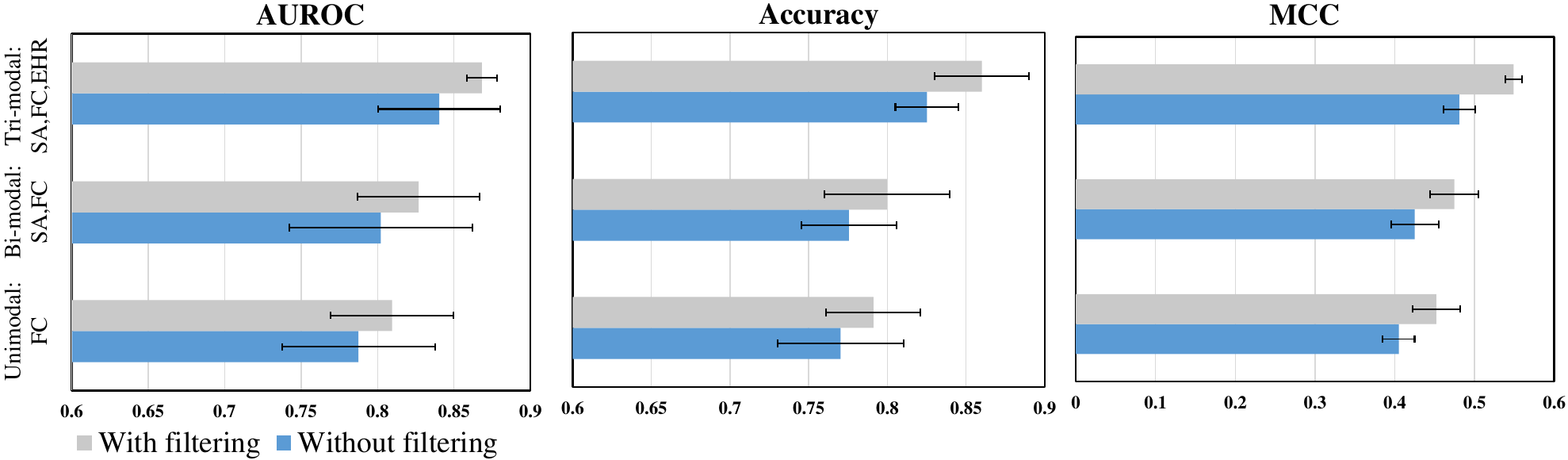}
		\caption{Effectiveness of uncertainty-based filtering.\label{fig:ablation1}}
	\end{subfigure}
	\medskip
	\begin{subfigure}[b]{0.48\textwidth}
		\centering
		\includegraphics[width=\textwidth, height=2.75cm]{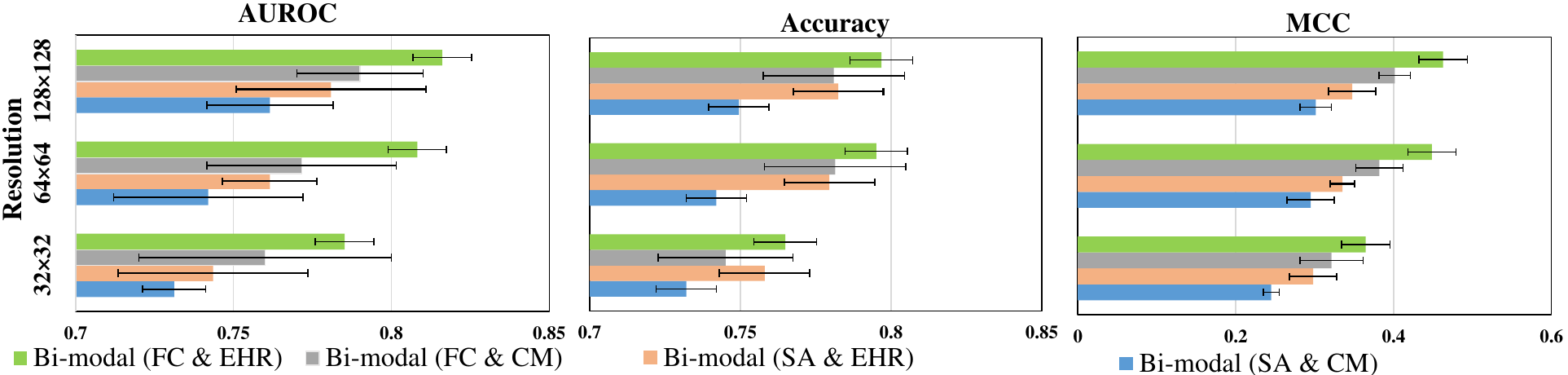}
		\caption{Importance of EHR features in bi-modal learning.\label{fig:ablation2}}
	\end{subfigure}
	\\
	\label{fig:three graphs}
	\centering
	\begin{subfigure}[b]{0.48\textwidth}
		\includegraphics[width=\textwidth, height=2.75cm]{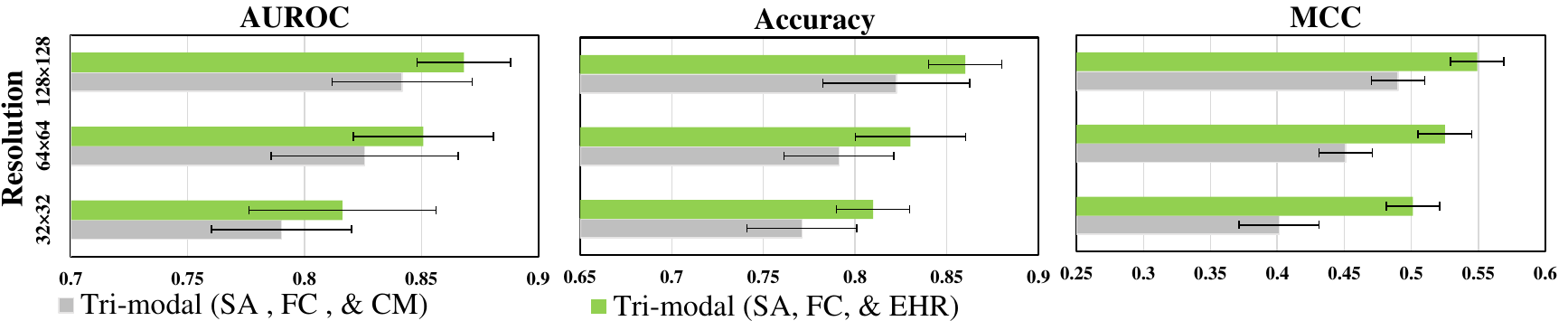}
		\caption{Importance of EHR features in tri-modal learning.\label{fig:ablation3}}
	\end{subfigure}
	\medskip
	\label{fig:four graphs}
	\centering
	\begin{subfigure}[b]{0.48\textwidth}
		\includegraphics[width=\textwidth, height=2.65cm]{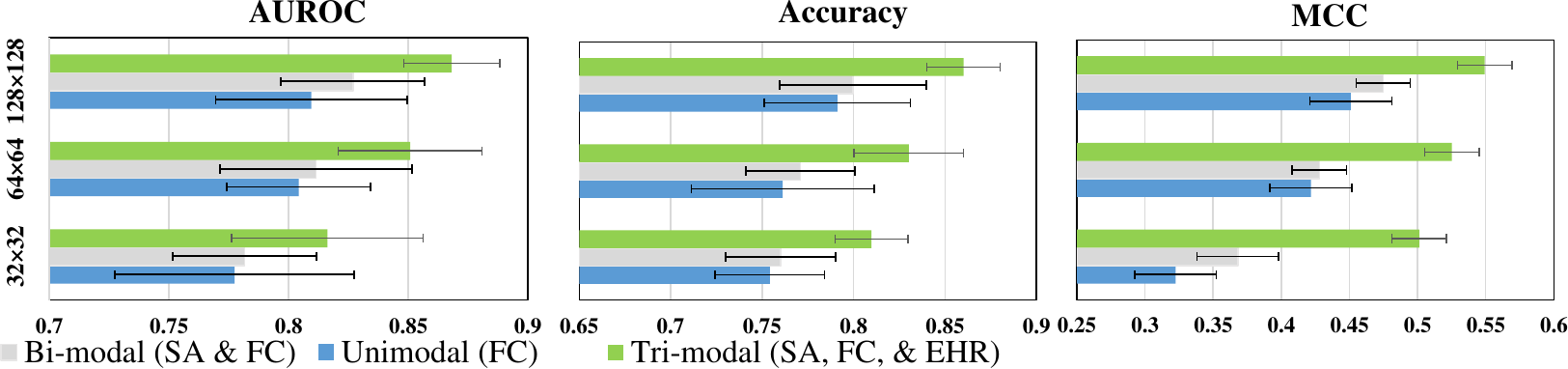}
		\caption{Tri-modal versus bi-modal and unimodal.}\label{fig:MUL}
	\end{subfigure}
	
	\caption{Ablation studies: (a) and (d) Unimodal (FC), Bi-modal (SA $\&$ FC), and Tri-modal (SA, FC, $\&$ EHR) are the best performing MPCA models in Tables~\ref{tab2},~\ref{tab3}, and~\ref{tab4}; (b) Bi-modal (FC $\&$ EHR), Bi-modal (FC $\&$ CM), Bi-modal (SA $\&$ EHR), and Bi-modal (SA $\&$ CM) are late fusion-based MPCA models; (c) Tri-modal (SA, FC, $\&$ EHR) and Tri-modal (SA, FC, $\&$ CM) are hybrid fusion-based MPCA models. }\label{m71}
\end{figure*}

\subsection{Uncertainity-based Filtering}
For quality control, we used uncertainty-based filtering on the training set. The training data was partitioned into $50$ bins. We removed one bin at a time in descending order of their uncertainties. We observed AUROC scores obtained from $10$-fold cross-validation on the training set. As depicted in Fig.~\ref{fig3}, the validation performance consistently improved when removing bins ($\leq4$) for all resolutions. Thus, we removed $146$ low quality samples from the training set of $1,849$. The remaining $1,703$ training samples were used to carry out various experiments in the remaining of this paper.

\subsection{Results of Unimodal Study}
We compared six unimodal models in this study. The results of this experiment are reported in Table~\ref{tab2}. The clinical baseline~\citep{garg2022cardiac} performs the prediction based on two Cardiac Measurement (CM) features: left atrial volume and left ventricle mass. We also created a unimodal baseline that uses the $15$ EHR features.  In Table~\ref{tab2}, MPCA (SA) and MPCA (FC) represent unimodal variants of our pipeline for short-axis and four-chamber modalities. These variants work on tensor-based features extracted from MPCA. We can notice from Table~\ref{tab2} that MPCA-based methods outperformed clinical baseline~\citep{garg2022cardiac}. The results demonstrate an improvement of $\Delta$AUROC $=0.080$, $\Delta$Accuracy $=0.051$, and $\Delta$MCC $=0.244$ over the clinical baseline obtained by FC-based unimodal, which indicates that MPCA-based features have a diagnostic value. Furthermore, we achieved an improvement of $\Delta$AUROC $=0.020$, $\Delta$Accuracy $=0.035$, and $\Delta$MCC $=0.109$ while using $15$ EHR features instead of only two CM features. EHR features also contain these two CM features. If we compare the performance of our pipeline with the transformer-based method then we can observe that the results are very competitive. Our pipeline produced better results in accuracy (i.e., $\Delta$Accuracy $=0.009$), whereas the transformer-based method achieved better results in AUROC and MCC metrics (i.e., $\Delta$AUROC $=0.003$ and $\Delta$MCC $=0.006$). However, our pipeline is fully interpretable in contrast with transformer-based method.

\subsection{Results of Bi-modal Study}
In this study, we compared the performance of nine bi-modal models. We reported the results in Table~\ref{tab3}. We compared the performance using three fusion approaches. For the transformer method, we implemented early and late fusion. In early fusion, we fed the concatenated tensor into the transformer architecture. In late fusion, we fused predictions made by separate networks. We did not use intermediate fusion in transformer models due to architectural design constraints. The results show that the intermediate fusion of the short-axis and four-chamber produced better results for MPCA-based models. The late fusion of short-axis and four-chamber outperformed early fusion approaches for transformer-based methods. Furthermore, we can notice from the results in Table~\ref{tab3} that the performance of our best model (MPCA: SA $\&$ FC) is better than the transformer method. 


\subsection{Results of Tri-modal Study}
We compared the performance of five tri-modal models in Table~\ref{tab4}. The performance achieved by MPCA (SA, FC, $\&$ EHR) with hybrid fusion is better than all the other models. This tri-modal model is a hybrid tri-modal that first combines short-axis and four-chamber using intermediate fusion, and then performs the late fusion with EHR features. MPCA (SA, FC, $\&$ EHR) with late fusion performs the second best among all comparing methods. These tri-modal models are also interpretable as they work on the linear fusion of multimodal data. Table~\ref{tab4} shows that our best MPCA-based tri-modal model achieved an improvement of $\Delta$AUROC $=0.011$, $\Delta$Accuracy $=0.016$, and $\Delta$MCC $=0.014$ over the best transformer-based tri-modal model. Furthermore, our tri-modal model outperformed the best bi-modal and uni-modal models with improvements of ($\Delta$AUROC $=0.041$, $\Delta$Accuracy $=0.060$, and $\Delta$MCC $=0.074$) and ($\Delta$AUROC $=0.056$, $\Delta$Accuracy $=0.077$, and $\Delta$MCC $=0.091$), respectively.

\subsection{Ablation Studies}
\subsubsection{Effectivness of Sample Filtering} 
We performed an ablation experiment to analyze the effectiveness of sample filtering using our best models. In this experiment, we chose MPCA (FC) for unimodal model, MPCA (SA $\&$ FC) with an intermediate fusion for bi-modal model, and MPCA (SA, FC, $\&$ EHR) with hybrid fusion for tri-modal model. We used the whole training set when we did not apply uncertainty-based filtering. Fig.~\ref{fig:ablation1} depicts the comparison results. The results show that the filtering enhances the performance for all metrics.

\subsubsection{Importance of EHR Features}
We studied the importance of EHR features in bi-modal and tri-modal learning. EHR features contain $15$ selected features that incorporate $11$ cardiac measurements and $4$ demographic features. The selected demographic features are age, gender, Body Surface Area (BSA), and ethnicity.  We compared the performance of our models when we used CM features based on clinical baseline~\citep{garg2022cardiac}. Fig.~\ref{fig:ablation2} shows the importance of EHR features on bi-modal learning. The results show that the models with EHR features outperformed other models. The importance of EHR features in tri-modal learning is shown in Fig.~\ref{fig:ablation3}. The results show that tri-modal model with EHR features produced better results than the tri-modal model with CM features. These results validates the importance of EHR features in our pipeline.

\subsubsection{Effectiveness of Multimodal Learning}
We studied the importance of multimodal learning in our pipeline. We compared our tri-modal against the unimodal and bi-modal models for three resoulutions in Fig.~\ref{fig:MUL}. We used the best MPCA models obtained from Table~\ref{tab2},~\ref{tab3}, and ~\ref{tab4} to perform this experiment. The bi-modal model produced better performance than the unimodal model and the tri-modal model outperformed all. This experiment shows that multimodal learning improves cardiovascular hemodynamics prediction.


\begin{figure}[t]
\centering
\includegraphics[scale=0.385]{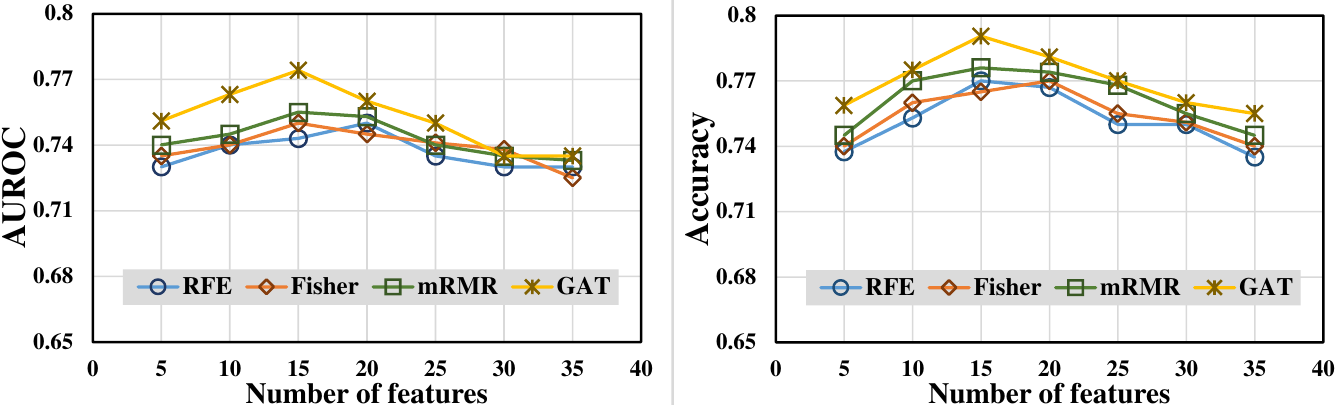}
\caption{Comparison of four feature selection methods on validation set for \textit{Baseline (EHR)} model.\label{fig:ablationGAT}}
\end{figure}

\begin{figure}[t]
\centering
\includegraphics[scale=0.26]{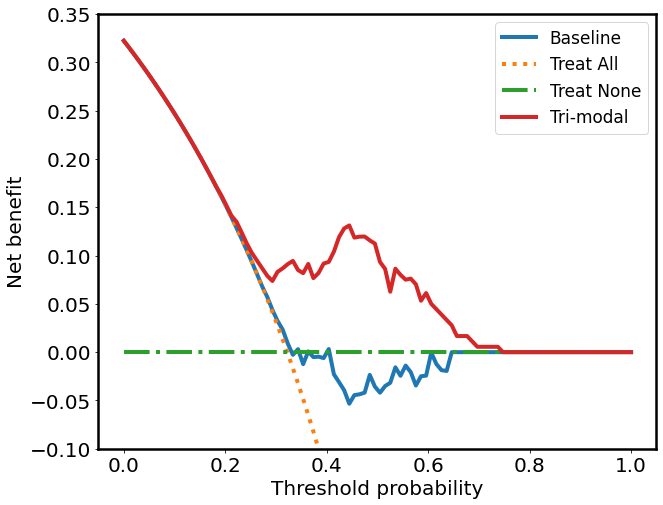}
\caption{Assessing the clinical utility of our tri-modal model using Decision Curve Analysis (DCA)~\cite{vickers2006decision}. ``Treat All" shows treating all patients, regardless of their actual disease status, whereas ``Treat None" shows treating no patients. Our model's net benefit is compared with the net benefit of treating everyone or no one to determine its overall utility.\label{fig:dca}}
\end{figure}

\begin{figure}[t]
\centering
\includegraphics[scale=0.33]{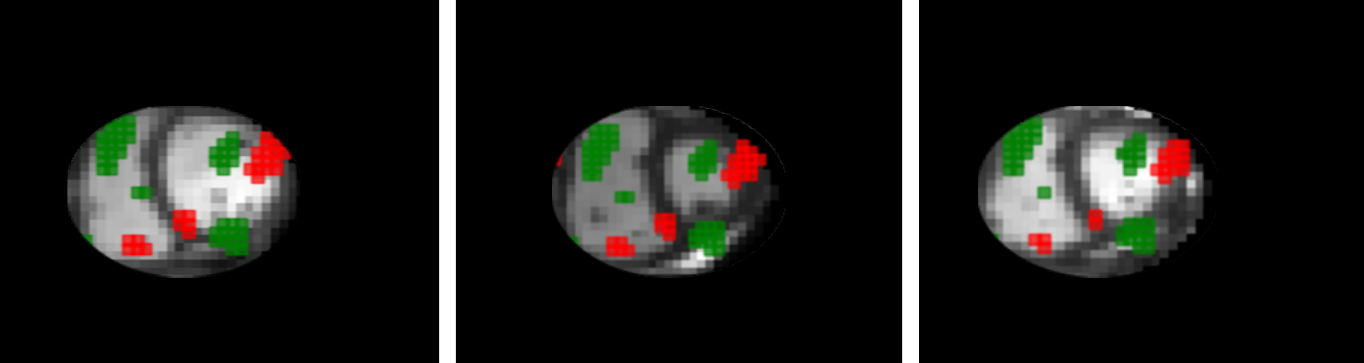}
\caption{Feature interpretation on cardiac scans of three subjects. Red: high-risk features, green: low-risk features.}
\label{interpret}
\end{figure}

\subsection{Effectiveness of GAT Feature Selection}
We selected $15$ EHR features from $49$ features using GAT feature selection. We evaluated the performance of GAT-based feature selection on different numbers of features in Fig~\ref{fig:ablationGAT} compared against three popular feature selection methods RFE, Fisher score, and mRMR. The results show that GAT outperformed other methods and achieved the best performance at $15$ EHR features.

\subsection{Decision Curve Analysis (DCA)}
Decision Curve Analysis (DCA)~\cite{vickers2006decision,sadatsafavi2021moving} assesses the clinical utility of a method at the population level. We carried out DCA on our best tri-modal model with the hybrid fusion of three modalities. Fig.~\ref{fig:dca} shows the result of DCA. Our tri-modal model outperformed the baseline method~\citep{garg2022cardiac} for most possible benefit preferences. The positive values indicate a net benefit (i.e., correct diagnosis), whereas the negative values show a net harm (i.e., incorrect diagnosis).
 Our tri-modal model achieved a promising net benefit between decision threshold probabilities of $0.30$ and $0.70$. This result confirms the clinical utility of our method for screening high-risk patients from a large population.

\subsection{Feature Interpretation}
The interpretation of features plays a crucial role in clinical diagnosis. Simple and transparent models are more easily accepted in clinical practice as they provide more understandable results. Clinicians often like to correlate the prediction of the model with the clinical criteria to gain confidence. We attempted to make our pipeline fully interpretable. The results produced for unimodal, bi-modal, and tri-modal variants of our pipeline are interpretable. We used a linear kernel in SVM implementation and our fusion strategies are also linear. We harnessed the capability of non-linear models for landmark localization and EHR feature selection but these tasks do not affect the interpretability of the prediction. The landmark localization is essentially a preprocessing step and the raw selected EHR features are used in our complete pipeline without compromising interpretability. Fig.~\ref{interpret} depicts the features identified on three CMR scans. We used two colors to mark the regions in the image. The red color shows high-risk features for raised PAWP, whereas the green color represents the regions for normal PAWP prediction. We observed high-risk features in the interventricular septum and left ventricle. In EHR features, cardiac measurement features were found more useful than demographic features. Among measurement features, left atrial volume and left ventricular mass contributed more than other features.  Age and Body Surface Area (BSA) were more valuable than other demographic features for the prediction.

\section{Conclusion}
\label{conclusion}
We proposed a multimodal-learning-based pipeline for cardiovascular hemodynamics prediction. We used three modalities including short-axis CMR, four-chamber CMR, and EHR data. We extracted the spatio-temporal features for CMRs using tensor-based learning. We leveraged uncertainty quantification to filter quality training samples. We designed a graph attention network to select important EHR features. We designed four feature fusion techniques. The proposed pipeline was tested on ASPIRE registry dataset of $2,641$ subjects. Extensive experimental results showed that the tri-modal model with hybrid fusion produced the best performance in comparison to others. Our pipeline provided fine interpretability of the results. The high-risk features detected by our pipeline were found clinically important. In the current study, we used single-site data. In the future, we would like to explore the applicability of our pipeline on multi-site datasets. We would also like to evaluate our pipeline on other cardiac hemodynamics tasks.

\bibliographystyle{IEEEtran}
\bibliography{sample.bib}

\begin{thebibliography}{10}
\providecommand{\url}[1]{#1}
\csname url@samestyle\endcsname
\providecommand{\newblock}{\relax}
\providecommand{\bibinfo}[2]{#2}
\providecommand{\BIBentrySTDinterwordspacing}{\spaceskip=0pt\relax}
\providecommand{\BIBentryALTinterwordstretchfactor}{4}
\providecommand{\BIBentryALTinterwordspacing}{\spaceskip=\fontdimen2\font plus
\BIBentryALTinterwordstretchfactor\fontdimen3\font minus \fontdimen4\font\relax}
\providecommand{\BIBforeignlanguage}[2]{{%
\expandafter\ifx\csname l@#1\endcsname\relax
\typeout{** WARNING: IEEEtran.bst: No hyphenation pattern has been}%
\typeout{** loaded for the language `#1'. Using the pattern for}%
\typeout{** the default language instead.}%
\else
\language=\csname l@#1\endcsname
\fi
#2}}
\providecommand{\BIBdecl}{\relax}
\BIBdecl

\bibitem{savarese2022global}
G.~Savarese, P.~M. Becher, L.~H. Lund, P.~Seferovic, G.~M. Rosano, and A.~J. Coats, ``Global burden of heart failure: a comprehensive and updated review of epidemiology,'' \emph{Cardiovascular Research}, vol. 118, no.~17, pp. 3272--3287, 2022.

\bibitem{emdin2009old}
M.~Emdin, S.~Vittorini, C.~Passino, and A.~Clerico, ``Old and new biomarkers of heart failure,'' \emph{Eur. J. Heart Fail.}, vol.~11, no.~4, pp. 331--335, 2009.

\bibitem{adamson2014wireless}
P.~B. Adamson, W.~T. Abraham, R.~C. Bourge, M.~R. Costanzo, A.~Hasan, C.~Yadav, J.~Henderson, P.~Cowart, and L.~W. Stevenson, ``Wireless pulmonary artery pressure monitoring guides management to reduce decompensation in heart failure with preserved ejection fraction,'' \emph{Circulation: Heart Failure}, vol.~7, no.~6, pp. 935--944, 2014.

\bibitem{garg2022cardiac}
P.~Garg, R.~Gosling, P.~Swoboda, R.~Jones, A.~Rothman, J.~M. Wild, D.~G. Kiely, R.~Condliffe, S.~Alabed, and A.~J. Swift, ``Cardiac magnetic resonance identifies raised left ventricular filling pressure: prognostic implications,'' \emph{Eur. Heart J.}, vol.~43, no.~26, pp. 2511--2522, 2022.

\bibitem{avendi2016combined}
M.~R. Avendi, A.~Kheradvar, and H.~Jafarkhani, ``A combined deep-learning and deformable-model approach to fully automatic segmentation of the left ventricle in cardiac {MRI},'' \emph{Med Image Anal .}, vol.~30, pp. 108--119, 2016.

\bibitem{assadi2022role}
H.~Assadi, S.~Alabed, A.~Maiter, M.~Salehi, R.~Li, D.~P. Ripley, R.~J. Van~der Geest, Y.~Zhong, L.~Zhong, A.~J. Swift \emph{et~al.}, ``The role of artificial intelligence in predicting outcomes by cardiovascular magnetic resonance: a comprehensive systematic review,'' \emph{Medicina}, vol.~58, no.~8, p. 1087, 2022.

\bibitem{fotaki2022artificial}
A.~Fotaki, E.~Puyol-Ant{\'o}n, A.~Chiribiri, R.~Botnar, K.~Pushparajah, and C.~Prieto, ``Artificial intelligence in cardiac {MRI}: is clinical adoption forthcoming?'' \emph{Front. Cardiovasc. Med.}, vol.~8, p. 818765, 2022.

\bibitem{alabed2022validation}
S.~Alabed, F.~Alandejani, K.~Dwivedi, K.~Karunasaagarar, M.~Sharkey, P.~Garg, P.~J. de~Koning, A.~T{\'o}th, Y.~Shahin, C.~Johns \emph{et~al.}, ``Validation of artificial intelligence cardiac {MRI} measurements: relationship to heart catheterization and mortality prediction,'' \emph{Radiology}, vol. 305, no.~1, pp. 68--79, 2022.

\bibitem{vimalesvaran2022detecting}
K.~Vimalesvaran, F.~Uslu, S.~Zaman, C.~Galazis, J.~Howard, G.~Cole, and A.~A. Bharath, ``Detecting aortic valve pathology from the 3-chamber cine cardiac {MRI} view,'' in \emph{Proc. MICCAI}, 2022, pp. 571--580.

\bibitem{priya2021radiomics}
S.~Priya, T.~Aggarwal, C.~Ward, G.~Bathla, M.~Jacob, A.~Gerke, E.~A. Hoffman, and P.~Nagpal, ``Radiomics side experiments and dafit approach in identifying pulmonary hypertension using cardiac mri derived radiomics based machine learning models,'' \emph{Scientific Reports}, vol.~11, no.~1, p. 12686, 2021.

\bibitem{alkhanfar2023non}
D.~Alkhanfar, K.~Dwivedi, F.~Alandejani, Y.~Shahin, S.~Alabed, C.~Johns, P.~Garg, A.~Thompson, A.~M. Rothman, A.~Hameed \emph{et~al.}, ``Non-invasive detection of severe {PH} in lung disease using {Magnetic Resonance Imaging},'' \emph{Front. Cardiovasc. Med.}, vol.~10, p. 1016994, 2023.

\bibitem{diller2020utility}
G.-P. Diller, J.~Vahle, R.~Radke, M.~L.~B. Vidal, A.~J. Fischer, U.~M. Bauer, S.~Sarikouch, F.~Berger, P.~Beerbaum, H.~Baumgartner \emph{et~al.}, ``Utility of deep learning networks for the generation of artificial cardiac magnetic resonance images in congenital heart disease,'' \emph{BMC Medical Imaging}, vol.~20, pp. 1--8, 2020.

\bibitem{swift2021machine}
A.~J. Swift, H.~Lu, J.~Uthoff, P.~Garg, M.~Cogliano, J.~Taylor, P.~Metherall, S.~Zhou, C.~S. Johns, S.~Alabed \emph{et~al.}, ``A machine learning cardiac magnetic resonance approach to extract disease features and automate pulmonary arterial hypertension diagnosis,'' \emph{Eur. Heart J. Cardiovasc. Imaging}, vol.~22, no.~2, pp. 236--245, 2021.

\bibitem{uthoff2020geodesically}
J.~Uthoff, S.~Alabed, A.~J. Swift, and H.~Lu, ``Geodesically smoothed tensor features for pulmonary hypertension prognosis using the heart and surrounding tissues,'' in \emph{Proc. MICCAI}, 2020, pp. 253--262.

\bibitem{alabed2022machine}
S.~Alabed, J.~Uthoff, S.~Zhou, P.~Garg, K.~Dwivedi, F.~Alandejani, R.~Gosling, L.~Schobs, M.~Brook, Y.~Shahin \emph{et~al.}, ``Machine learning cardiac-{MRI} features predict mortality in newly diagnosed pulmonary arterial hypertension,'' \emph{European Heart Journal-Digital Health}, vol.~3, no.~2, pp. 265--275, 2022.

\bibitem{curiale2019automatic}
A.~H. Curiale, F.~D. Colavecchia, and G.~Mato, ``Automatic quantification of the lv function and mass: A deep learning approach for cardiovascular {MRI},'' \emph{Comput Methods Programs Biomed}, vol. 169, pp. 37--50, 2019.

\bibitem{gosling2023quantifying}
R.~C. Gosling, G.~Williams, A.~Al~Baraikan \emph{et~al.}, ``Quantifying myocardial blood flow and resistance using 4d-flow cardiac magnetic resonance imaging,'' \emph{Cardiology Research and Practice}, vol. 2023, 2023.

\bibitem{puyol2020interpretable}
E.~Puyol-Ant{\'o}n, C.~Chen, J.~R. Clough \emph{et~al.}, ``Interpretable deep models for cardiac resynchronisation therapy response prediction,'' in \emph{Proc. MICCAI}, 2020, pp. 284--293.

\bibitem{ahmadi2023transformer}
N.~Ahmadi, M.~Tsang, A.~Gu, T.~Tsang, and P.~Abolmaesumi, ``Transformer-based spatio-temporal analysis for classification of aortic stenosis severity from echocardiography cine series,'' \emph{IEEE Trans. Med. Imaging}, 2023.

\bibitem{schobs2022uncertainty}
L.~A. Schobs, A.~J. Swift, and H.~Lu, ``Uncertainty estimation for heatmap-based landmark localization,'' \emph{IEEE Trans. Med. Imaging}, vol.~42, no.~4, pp. 1021--1034, 2022.

\bibitem{lu2008mpca}
H.~Lu, K.~N. Plataniotis, and A.~N. Venetsanopoulos, ``{MPCA}: Multilinear principal component analysis of tensor objects,'' \emph{IEEE Trans. on Neural Netw.}, vol.~19, no.~1, pp. 18--39, 2008.

\bibitem{soenksen2022integrated}
L.~R. Soenksen, Y.~Ma, C.~Zeng, L.~Boussioux, K.~Villalobos~Carballo, L.~Na, H.~M. Wiberg, M.~L. Li, I.~Fuentes, and D.~Bertsimas, ``Integrated multimodal artificial intelligence framework for healthcare applications,'' \emph{NPJ Digital Medicine}, vol.~5, no.~1, p. 149, 2022.

\bibitem{10.1007/978-3-031-43990-2_20}
P.~C. Tripathi, M.~N.~I. Suvon, L.~Schobs, S.~Zhou, S.~Alabed, A.~J. Swift, and H.~Lu, ``Tensor-based multimodal learning for prediction of pulmonary arterial wedge pressure from cardiac {MRI},'' in \emph{Proc. MICCAI}, 2023, pp. 206--215.

\bibitem{welch1947generalization}
B.~L. Welch, ``The generalization of ‘student's’problem when several different population varlances are involved,'' \emph{Biometrika}, vol.~34, no. 1-2, pp. 28--35, 1947.

\bibitem{hurdman2012aspire}
J.~Hurdman, R.~Condliffe, C.~Elliot \emph{et~al.}, ``{ASPIRE} registry: assessing the {Spectrum of Pulmonary hypertension Identified at a REferral centre},'' \emph{Eur. Respir. J.}, vol.~39, no.~4, pp. 945--955, 2012.

\bibitem{lu2013multilinear}
H.~Lu, K.~N. Plataniotis, and A.~Venetsanopoulos, \emph{Multilinear subspace learning: dimensionality reduction of multidimensional data}.\hskip 1em plus 0.5em minus 0.4em\relax CRC press, 2013.

\bibitem{li2017feature}
J.~Li, K.~Cheng, S.~Wang, F.~Morstatter, R.~P. Trevino, J.~Tang, and H.~Liu, ``Feature selection: A data perspective,'' \emph{ACM Computing Surveys}, vol.~50, no.~6, pp. 1--45, 2017.

\bibitem{wang2021mogonet}
T.~Wang, W.~Shao, Z.~Huang, H.~Tang, J.~Zhang, Z.~Ding, and K.~Huang, ``Mogonet integrates multi-omics data using graph convolutional networks allowing patient classification and biomarker identification,'' \emph{Nature Communications}, vol.~12, no.~1, p. 3445, 2021.

\bibitem{veličković2018graph}
P.~Veličković, G.~Cucurull, A.~Casanova, A.~Romero, P.~Liò, and Y.~Bengio, ``Graph attention networks,'' in \emph{Proc. ICLR}, 2018.

\bibitem{hamilton2020graph}
W.~L. Hamilton, ``Graph representation learning,'' \emph{Synthesis Lectures on Artificial Intelligence and Machine Learning}, vol.~14, no.~3, pp. 1--159, 2020.

\bibitem{setiono1997neural}
R.~Setiono and H.~Liu, ``Neural-network feature selector,'' \emph{IEEE Trans. on Neural Netw.}, vol.~8, no.~3, pp. 654--662, 1997.

\bibitem{9468895}
R.~A. Amjad, K.~Liu, and B.~C. Geiger, ``Understanding neural networks and individual neuron importance via information-ordered cumulative ablation,'' \emph{IEEE Trans. Neural Netw. Learn. Syst.}, vol.~33, no.~12, pp. 7842--7852, 2022.

\bibitem{JMLR:v15:srivastava14a}
N.~Srivastava, G.~Hinton, A.~Krizhevsky, I.~Sutskever, and R.~Salakhutdinov, ``Dropout: A simple way to prevent neural networks from overfitting,'' \emph{J. Mach. Learn. Res.}, vol.~15, no.~56, pp. 1929--1958, 2014.

\bibitem{FeyLenssen2019}
M.~Fey and J.~E. Lenssen, ``Fast graph representation learning with {PyTorch Geometric},'' in \emph{Proc. ICLR Workshop on Representation Learning on Graphs and Manifolds}, 2019.

\bibitem{kingma2015adam}
D.~P. Kingma and J.~Ba, ``Adam: A method for stochastic optimization,'' in \emph{Proc. ICLR}, 2015.

\bibitem{chicco2020advantages}
D.~Chicco and G.~Jurman, ``The advantages of the {Matthews Correlation Coefficient} ({MCC}) over {F1} score and accuracy in binary classification evaluation,'' \emph{BMC Genomics}, vol.~21, no.~1, pp. 1--13, 2020.

\bibitem{lu2022pykale}
H.~Lu, X.~Liu, S.~Zhou, R.~Turner, P.~Bai, R.~E. Koot, M.~Chasmai, L.~Schobs, and H.~Xu, ``Pykale: Knowledge-aware machine learning from multiple sources in python,'' in \emph{Proc. CIKM}, 2022, pp. 4274--4278.

\bibitem{pedregosa2011scikit}
F.~Pedregosa, G.~Varoquaux, A.~Gramfort \emph{et~al.}, ``Scikit-learn: Machine learning in python,'' \emph{J. Mach. Learn. Res.}, vol.~12, pp. 2825--2830, 2011.

\bibitem{10.5555/954544}
R.~O. Duda, P.~E. Hart, and D.~G. Stork, \emph{Pattern Classification}.\hskip 1em plus 0.5em minus 0.4em\relax Wiley-Interscience, 2000.

\bibitem{mrmr2025peng}
H.~Peng, F.~Long, and C.~Ding, ``Feature selection based on mutual information criteria of max-dependency, max-relevance, and min-redundancy,'' \emph{IEEE Trans. Pattern Anal. Mach. Intell.}, vol.~27, no.~8, pp. 1226--1238, 2005.

\bibitem{li2018feature}
J.~Li, K.~Cheng, S.~Wang, F.~Morstatter, R.~P. Trevino, J.~Tang, and H.~Liu, ``Feature selection: A data perspective,'' \emph{ACM Computing Surveys}, vol.~50, no.~6, p.~94, 2018.

\bibitem{guyon2002gene}
I.~Guyon, J.~Weston, S.~Barnhill, and V.~Vapnik, ``Gene selection for cancer classification using {Support Vector Machines},'' \emph{Machine Learning}, vol.~46, pp. 389--422, 2002.

\bibitem{vickers2006decision}
A.~J. Vickers and E.~B. Elkin, ``Decision curve analysis: a novel method for evaluating prediction models,'' \emph{Medical Decision Making}, vol.~26, no.~6, pp. 565--574, 2006.

\bibitem{sadatsafavi2021moving}
M.~Sadatsafavi, A.~Adibi, M.~Puhan, A.~Gershon, S.~D. Aaron, and D.~D. Sin, ``Moving beyond {AUC}: decision curve analysis for quantifying net benefit of risk prediction models,'' \emph{Eur. Respir. J.}, vol.~58, no.~5, 2021.

\end{thebibliography}

\end{document}